\documentclass[twoside,11pt]{article}
\usepackage{todonotes}
\usepackage{blindtext}
\usepackage{caption}
\usepackage{subcaption}
\usepackage{makecell}
\usepackage{amsmath}
\usepackage{booktabs}
\usepackage{subcaption}
\usepackage{float}
\usepackage[numbers]{natbib}

\usepackage[abbrvbib, preprint]{jmlr2e}

\ShortHeadings{Multi-Modal Signal Classification}{Kuppel et al.}
\firstpageno{1}

\begin{document}

\title{Multi-modal transformer for signal classification in
nanopore blockade experiments}

\author{\name Sandro Kuppel$^{\dagger}$ \email skuppel@icp.uni-stuttgart.de \\
       \addr Institute for Computational Physics\\
       University of Stuttgart\\
       Allmandring 3, 70569 Stuttgart, Germany
       \AND
       \name Julian Hoßbach$^{\dagger}$ \email jhossbach@icp.uni-stuttgart.de \\
       \addr Institute for Computational Physics\\
       University of Stuttgart\\
       Allmandring 3, 70569 Stuttgart, Germany
       \AND
       \name Samuel Tovey \email stovey@icp.uni-stuttgart.de \\
       \addr Institute for Computational Physics\\
       University of Stuttgart\\
       Allmandring 3, 70569 Stuttgart, Germany
       \AND
       \name Christian Holm$^{\ast}$ \email holm@icp.uni-stuttgart.de \\
       \addr Institute for Computational Physics\\
       University of Stuttgart\\
       Allmandring 3, 70569 Stuttgart, Germany\\
       \AND
       \small$^\dagger$These authors contributed equally to this work.\\
       \small$^\ast$Corresponding author. Email: holm@icp.uni-stuttgart.de
       }


\maketitle

\begin{abstract}
Nanopore devices have emerged as powerful tools for single-molecule sensing, with potential for rapid, portable diagnostics. 
They detect changes in ionic current as analytes enter nanometer-scale pores, providing a means of identifying diverse biomarkers from their characteristic signal patterns. 
However, these signals are highly complex, and reliably assigning them to specific molecules remains a major challenge. 
Here, we address this by introducing a multi-modal deep learning architecture that jointly processes multiple signal representations, including raw time-series data, wavelet-based images, and static feature vectors. 
Our approach surpasses existing methods by more than 10 percentage points on a 42-peptide benchmark and transfers to a 20-amino-acid dataset with near-perfect accuracy.
The model integrates complementary information from these representations, with attention analysis showing that the time-series and wavelet-image inputs emphasize different features of the same event.
Together, these results demonstrate the potential of machine learning to enable robust, high-accuracy molecular identification with nanopore sensors.

\end{abstract}


\section{Introduction}
Proteins play essential roles in our bodies, from transporting molecules and catalyzing biochemical processes to defending against pathogens.
Beyond the $\sim 20000$ protein-coding genes found by the Human Genome Project~\cite{internationalhumangenomesequencingconsortium04a}, this abundance of diverse functions is achieved through various alterations of the proteins, such as post-translational modifications (PTMs) of the proteins after their biosynthesis, or alternative splicing.
The ability to identify these proteins and their modifications requires sequencing and identification techniques with sensitivity to single-molecules and even functional groups, which is challenging to achieve with conventional methods~\cite{wei23a}.
Biological nanopores have emerged as promising candidates for single-molecule protein sensing capable of this sensitivity~\cite{alfaro21a,ratinho25a,motone2024multipass}.
In this approach, a biological nanopore is inserted into an insulating membrane that divides the cis and trans sides of a container filled with a concentrated aqueous salt solution. When a small potential difference is applied, an ionic current flowing through the pore is recorded.
When an analyte, such as a protein or peptide, enters the pore, the flow of ions is disrupted, and a drop in the current can be measured.\\
The prevalent method for performing this classification is to select fixed features from these blockade events.
A notable feature set is the combination of the analyte's dwell time in the pore and the mean blockade current it produces.
By computing histograms of these features after many samples have passed through a nanopore, the resulting distributions can be used to classify peptides and proteins with significant differences in their dwell time and blockade current values~\cite{ouldali20a,bakshloo22a,ensslen22a,ensslen24a, srnko2026histone}.
However, detecting analytes with subtle differences in their amino acid composition or PTMs remains challenging to achieve based solely on dwell time and mean blockade current, despite numerous advances in experimental techniques designed to enhance sensitivity and resolution~\cite{ratinho25a}.
Similar arguments can be made about other more sophisticated static feature sets used in the literature classification task~\cite{zhang24a,yu23a,hossbachPeptideClassificationStatistical2024}.\\
A central challenge with static features is that they ignore temporal correlations within the data~\cite{yu23a,rukes25a-pre}.
The evolution of the blockade current is dependent on the correlated motion of the analyte in the pore.
Because the dynamics of an analyte is related to its biochemical interactions and structure inside the pore, a time-based correlation of the current signals should capture information relevant to the classification task.
Previous works involving machine learning approaches have utilized convolutional neural networks and vision transformers to capture these correlations, either directly on the raw ionic current trace or through wavelet transformed representations \cite{wang25a,misiunas18b,tovey25b-pre, shah2026interpretable, hart2026nanoboost}.
Although promising, these approaches focus on only a single representation of the data, which may not be optimal for extracting the most relevant features.\\
In this work, we address this limitation by combining information from different data representations in a deep learning model capable of processing them simultaneously.
As an additional baseline, we evaluated a transformer architecture that directly captures temporal correlations and signal characteristics from the raw time-series signals. 
We find that this and other single-modality methods exhibit varying performance across distinct peptide classes, indicating that each representation captures complementary aspects of the signal.
Our multi-modal model effectively integrates this complementary information and achieves a substantial improvement in classification accuracy, outperforming previous approaches and approaching what we surmise may be the limit in classification accuracy, based on the estimated label noise in the data.

\section{Results}
\begin{figure}[H]
	\centering
	\includegraphics[width=0.7\linewidth]{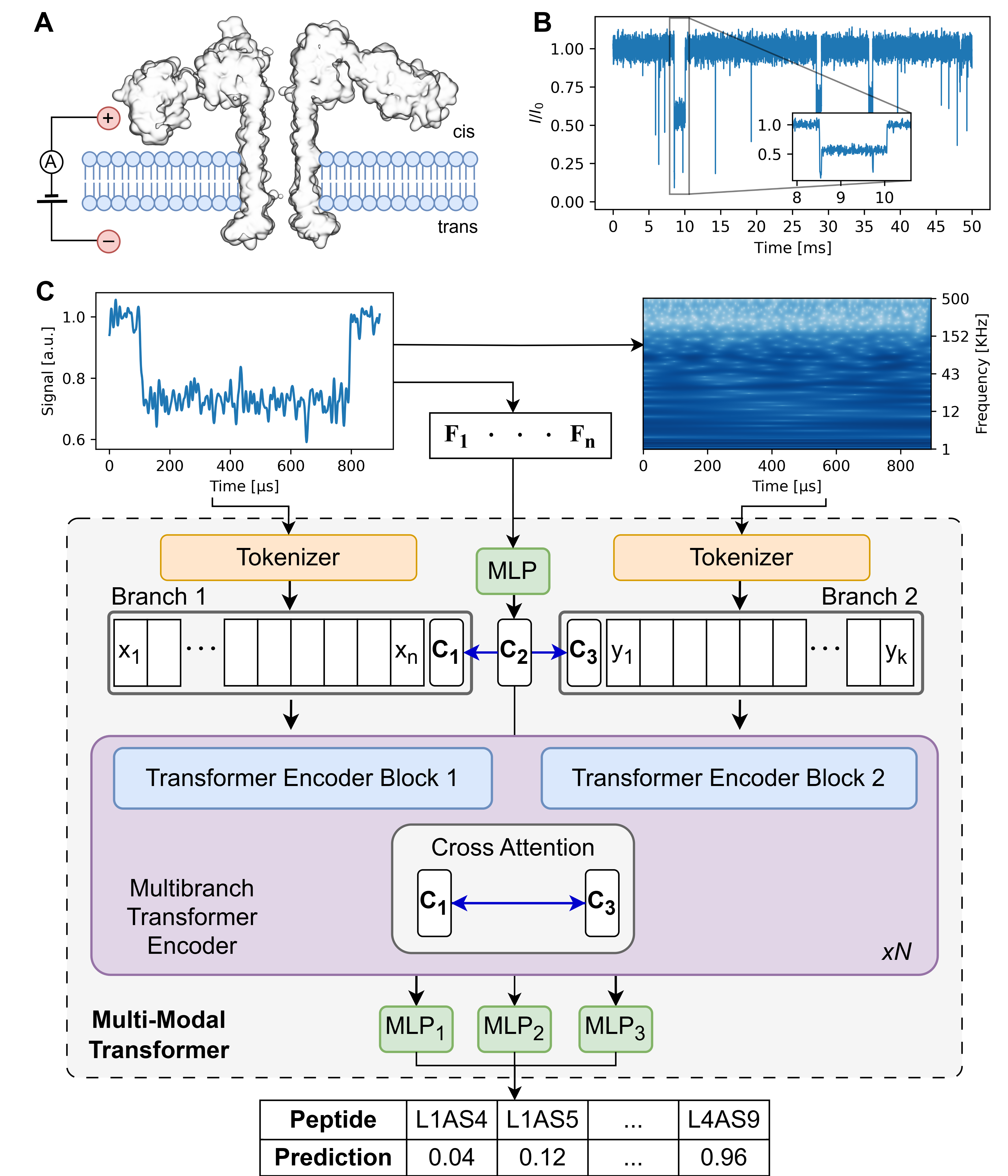}
	\caption{\textbf{Experimental workflow and model architecture.}
		\textbf{(A)} Illustration of the experimental setup using an aerolysin nanopore embedded in a lipid bilayer over which an electric current is applied.
		\textbf{(B)} Measuring the electric current through the pore results in a continuous current readout with distinctive blockade currents, corresponding to peptides being present in the pore.
		These events are extracted and analyzed.
		\textbf{(C)} For classification, each current trace is converted into an image via wavelet transformation, and a feature vector is computed. These three modalities, the raw trace, wavelet image, and descriptor vector, serve as inputs to our multi-modal transformer.
		All modalities are first projected into the model's latent space. Time-series and image data are divided into equal-length segments and square patches, respectively, then passed through linear layers to form input tokens $x_{1}\ldots x_{n}$ and $y_{1}\ldots y_{k}$, with added 
        positional embeddings and learnable classification tokens $C_{1}$ and $C_{3}$ appended.
		The descriptor features $F_{1}\ldots F_{m}$ are processed by a multilayer perceptron (MLP) to produce the classification token $C_{2}$. Cross-attention (blue arrows) transfers information once from $C_{2}$ to $C_{1}$ and $C_{3}$.
		Branches one and two are then processed by a multi-branch transformer encoder, alternating standard transformer blocks and cross-attention for $N$ iterations.
		Finally, all classification tokens pass through separate MLPs and are combined to form the final classification vector.
	}
	\label{fig:architecture}
\end{figure}

In this study, we introduce a multi-modal transformer that classifies peptides directly from nanopore blockade events by jointly processing several representations of the same signal within a single model.
An overview of the complete workflow, including the experimental setup, data preprocessing, and model architecture, is provided in Fig.~\ref{fig:architecture}.
Starting from the ionic current traces recorded using a biological nanopore (Fig.~\ref{fig:architecture}A), we extract signal segments corresponding to individual peptides present in the pore (Fig.~\ref{fig:architecture}B).
Each extracted event is then transformed into two additional modalities (Fig.~\ref{fig:architecture}C): an image representation obtained by a wavelet transformation and a feature vector computed from the catch22 statistical descriptors~\cite{lubbaCatch22CAnonicalTimeseries2019}.
The three inputs are then simultaneously processed in the multi-modal transformer architecture shown in Fig.~\ref{fig:architecture}C.
The model is based on the Vision Transformer (ViT) framework~\cite{ImageWorth16x16}, which we have extended to handle multiple modalities, similar to the approach in~\cite{chenCrossViTCrossAttentionMultiScale2021}.
Each input type is processed in its own branch, while cross-attention enables communication between branches, allowing efficient fusion of information with minimal computational overhead.
Although we use one input for each modality type, time series, images, and feature sets, the architecture can be easily extended to any number of inputs from these modality types.
In addition to the multi-modal model, we report a transformer trained on the time-series representation alone as an additional baseline.

\subsection*{Classification results}
To evaluate our proposed model, we train it on two datasets: The first dataset contains blockade events from 42 different peptides obtained through a peptide ladder experiment\cite{behrends22a,hossbachPeptideClassificationStatistical2024,tovey25b-pre}.
The data set contains 350,000 samples that are unevenly distributed between the different peptides, posing additional challenges for training and evaluation.
We compare the performance with two previously applied methods that use wavelet images \citep{tovey25b-pre} and the $catch22$ feature set ~\citep{hossbachPeptideClassificationStatistical2024} as single modalities, as well as the additional time series transformer baseline.
Detailed information on the data set, model design, and training procedures can be found in Section~\ref{sec:methods_data}.
Table~\ref{tab:acc_peptide} outlines the final macro-averaged, micro-averaged, maximum, and minimum accuracies for each model.
The macro-averaged accuracy is the average of the per-class accuracy, whereas the micro-averaged accuracy corresponds to the overall accuracy across all events.
Finally, the maximum and minimum accuracies correspond to the best and worst-performing classes, respectively.

\noindent The benefits of combining multiple modalities are evident in the results of our multi-modal transformer, as highlighted by the bold numbers in Table~\ref{tab:acc_peptide}.
Micro- and macro-averaged accuracies of 92.1\% and 92.6\% are achieved, resulting in an increase of more than 10 percentage points across both metrics compared to the best single-modality model.
Furthermore, the accuracy of the worst-performing class increased by almost 20 percentage points compared to the best-performing baseline, and the best-performing class is now classified with an accuracy of $100\%$.
Improving the worst-performing class is especially relevant for diagnostic applications, where a single poorly resolved analyte can compromise the reliability of the entire assay.

\begin{table}[h]
	\centering
	\caption{\textbf{Test accuracies.}
		Performance of our multi-modal and time series model, along with a comparison to prior work.}
	\begin{tabular}{lccccc}
		\toprule
		Model                                                         & Macro           & Micro           & Max            & Min             \\
		\midrule
		ResNet18 -- Wavelets \citep{tovey25b-pre}                                 & 81.7\%          & 81.5\%          & 95.6\%         & 58.7\%          \\
		MLP -- Catch22 ~\citep{hossbachPeptideClassificationStatistical2024} & 73.6\%          & 73.6\%          & 96.5\%         & 18.2\%          \\
		Time Series Transformer                                       & 77.0\%          & 79.6\%          & 92.8\%         & 25.5\%          \\
		Multi-Modal Transformer                                       & \textbf{92.6\%} & \textbf{92.1\%} & \textbf{100\%} & \textbf{77.2\%} \\
		\bottomrule
	\end{tabular}
	\label{tab:acc_peptide}
\end{table}

\begin{figure}[h]
	\centering
	\includegraphics[width=0.99\linewidth]{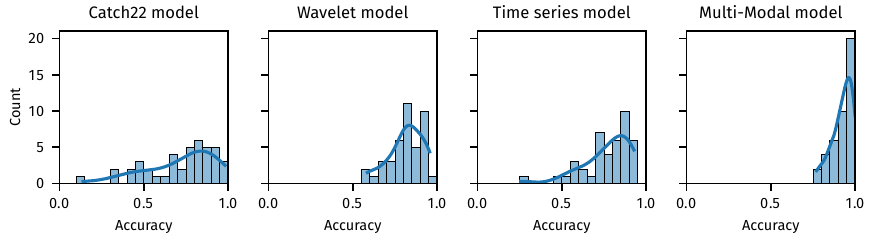}
	\caption{\textbf{Per-class accuracy distributions on the 42-peptide ladder dataset.} Distribution of per-class accuracies for all four models -- ResNet18, catch22, the time-series transformer, and the multi-modal transformer -- evaluated on the 42-peptide ladder dataset. The multi-modal transformer concentrates near perfect accuracy, whereas the single-modality models show broader distributions with a tail of poorly classified classes.}
	\label{fig:acc-hist}
\end{figure}

\noindent To compare the models in more detail, we examine the distribution of per-class accuracies across all four models (Fig.~\ref{fig:acc-hist}).
The distributions reveal a clear progression in robustness across models.
The single-modality models show broad distributions with a tail of poorly classified classes: catch22 spans nearly the full accuracy range, while the wavelet and time-series representations perform better, but still leave a subset of classes at noticeably lower accuracy.
In contrast, the multi-modal transformer concentrates sharply above 0.75, classifying more than 15 of the 42 classes at close to perfect accuracy.
This narrow, high distribution highlights the robustness of the multi-modal model and pushes the accuracy distribution significantly closer to real-world applicability.

\begin{figure}[h]
	\centering
	\includegraphics[width=\linewidth]{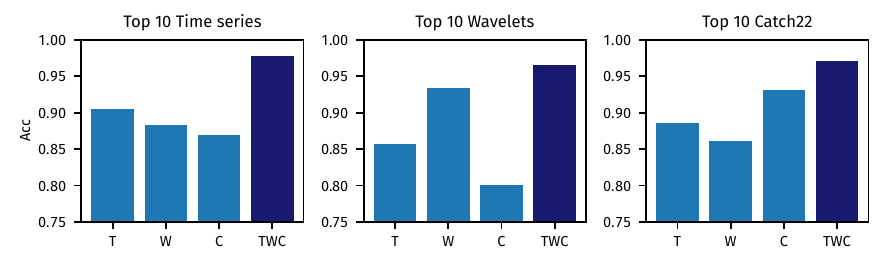}
	\caption{
		\textbf{Fusion evaluation.} To evaluate our model's ability to integrate information from all three modalities, we first identify the classes where each modality performs best.
		Specifically, we select the top 10 classes with the highest accuracies for each modality using their respective best-performing model.
		For the $catch22$ data, we use the dense neural network from~\cite{hossbachPeptideClassificationStatistical2024}, for the wavelet data, we show the ResNet18 from ~\cite{tovey25b-pre}, and for the raw time series, we show our transformer model.
		After identifying the top-performing classes for each modality, we calculate the average accuracies of these classes across the other modalities and our multi-modal approach (dark blue). The labels C, W, and T correspond to $catch22$, wavelet, and raw time series data, respectively.
	}
	\label{fig:fusion}
\end{figure}

\noindent Furthermore, we evaluated our model's ability to incorporate all relevant information from the three modalities.
To do so, we first identify the classes that are best recognized by each modality.
Specifically, we select the top 10 classes with the highest accuracies for each modality using their respective best-performing models.
For the $catch22$ data, we use the dense neural network from~\cite{hossbachPeptideClassificationStatistical2024}, for the wavelet data, we use the ResNet18 from ~\cite{tovey25b-pre}, and for the time series data, we use our Time Series Transformer.
After identifying the top-performing classes for each modality, we calculate the average accuracies of these classes across all models.
The results are presented in Fig.~\ref{fig:fusion}.
Examining the accuracies of the individual modalities (C=$catch22$, W=wavelets, and T=time series), it is evident that each modality performs particularly well in its respective top-performing classes.
Although the wavelet approach outperforms the other approaches in terms of micro- and macro-averaged accuracy, it is still beaten by the specific top-performing classes of the other modalities.
This observation indicates that each model extracts unique information from the respective representations on which it is trained.
The multi-modal model (TWC), shown in dark blue, not only matches but surpasses the performance of the individual models across all three cases.
These results demonstrate that our model successfully integrates relevant information from all modalities and that the combination further enhances the classification performance.

\subsection*{Model analysis}
\begin{figure}[h]
    \centering
    \includegraphics[width=0.7\linewidth]{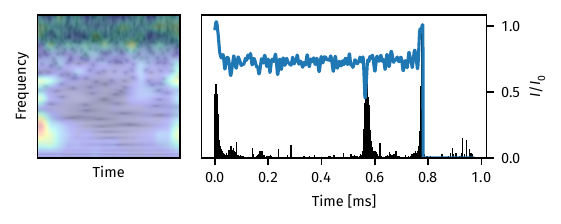}
    \caption{\textbf{Attention analysis of a representative blockade event.}
    Attention rollout~\citep{abnar_quantifying_2020} for the Time Series Transformer (left) and the Vision Transformer (right), shown for the same event in its raw time-series and wavelet representation.
    For the time series, attention weights are displayed as black bars beneath the current trace; the zero padding applied to reach a uniform input length is not masked and is therefore shown as well.
    For the wavelet image, attention is overlaid on the grayscale transform, with red indicating the highest values.
    The time-series model attends to the entry and exit phases and to the deep blockade in the centre of the event, whereas the wavelet model attends to the entry and exit phases and to the high-frequency band throughout.}
    \label{fig:attention}
\end{figure}

The preceding results show that models trained on different representations perform differently across peptide classes and that combining them yields a clear improvement in classification accuracy.
To understand the origin of this complementarity, we ask what information each representation contributes on its own, and analyze attention in the Vision Transformer and the Time Series Transformer, which are trained on a single representation each.
We apply attention rollout~\citep{abnar_quantifying_2020}, which aggregates attention across all layers and heads to give a global view of the information flow.
\autoref{fig:attention} shows the resulting attention for one representative event, in both its time-series and wavelet representation.\\
\noindent In the time-series representation, attention concentrates on the entry and exit phases of the event and on a pronounced deep blockade in the center of the signal.
Such deep and short blockades have been attributed to a binding site near the cis opening of the pore, where a positively charged physico-electric barrier and reduced volume are located~\citep{ensslen24b, boukhet18a}, suggesting that peptide interactions with this site carry discriminatory information.
A smaller amount of attention falls on the padded region beyond the signal, which may indicate that the model uses the empty region as a reference frame to infer dwell time, or as a register to correlate information extracted from the signal itself, as observed in~\citep{darcet24a-pre}.\\
\noindent The wavelet representation highlights the same entry and exit phases at the edges of the image but distributes its remaining attention differently: it attends strongly across the high-frequency band throughout the event and only weakly to the time position of the deep blockade that dominates the time-series map.
This indicates that the two representations expose different aspects of the same event.
The emphasis on high-frequency components is consistent with previous findings that these carry meaningful information rather than pure noise~\citep{tovey25b-pre}.
We hypothesize that such high-frequency information, made explicit by the wavelet transform, is harder for the model to extract directly from the raw trace, offering a possible explanation for the improved performance obtained when both representations are combined.\\
\noindent These patterns are not specific to the event shown but are consistently observed across randomly selected signals.
Further examples, together with final-layer attention maps resolved by individual attention heads, are provided in Appendix~\ref{app:AttAnal}.

\subsection*{Transfer learning}

\begin{figure}[h]
    \centering
    \includegraphics[width=0.99\linewidth]{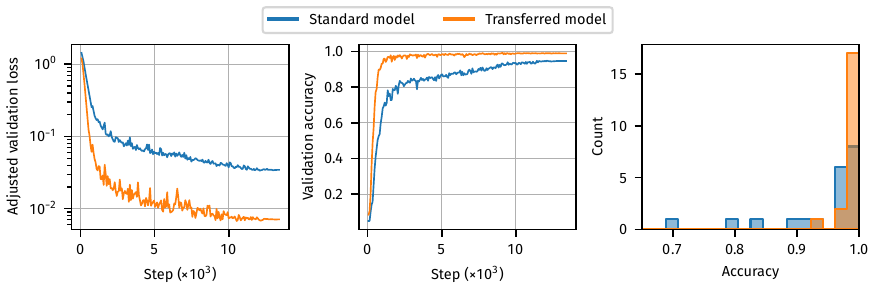}
    \caption{\textbf{Transfer learning evaluation.} Validation loss and accuracy during training (left) for a model trained on the $\mathrm{X_{R7}}$ dataset alone (standard) and one pre-trained on the peptide-ladder dataset and fine-tuned on $\mathrm{X_{R7}}$ (transferred). Because label smoothing prevents the cross-entropy loss from reaching zero even for a perfect classifier, we subtract this irreducible offset and report the adjusted loss. Per-class accuracy distributions of both final models (right). The transferred model converges faster and shows a narrower accuracy distribution.}
    \label{fig:trans-learning}
\end{figure}

Finally, we investigate the transfer-learning capabilities of our multi-modal model.
Data in protein nanopore experiments are often scarce, which limits both the application of machine learning methods and their early testing before large datasets have been generated.
A foundation model that has already learned to extract broadly relevant features from nanopore signals would mitigate this limitation.
To test whether our multi-modal model can serve this role, we train on the $\mathrm{X_{R7}}$ dataset of \cite{ouldali20a}, containing residual current events of all 20 proteinogenic amino acids linked to arginine heptapeptide carriers (see Section~\ref{sec:methods_data}), in two ways: once following the same training setup used for the peptide-ladder dataset and once by fine-tuning a model that had been fully trained on the peptide-ladder dataset.
With 38,000 labeled events, the $\mathrm{X_{R7}}$ dataset is roughly an order of magnitude smaller than the peptide-ladder dataset, placing it in the low-data regime that motivates pre-training.
However, the classification task is also slightly easier, comprising only 20 classes instead of 42, of which a subset is separable by the mean residual current alone~\citep{ouldali20a}, as is reflected in the higher accuracies both models reach.
The results are shown in \autoref{fig:trans-learning}.
The transferred model learns substantially faster and reaches a slightly higher final accuracy and lower final loss, indicating that training on the peptide-ladder dataset yields features that generalize beyond it.
The accuracy distributions support this. The transferred model shows a markedly narrower spread, with its worst-performing class still exceeding 90\%, whereas the standard model falls as low as 70\%.
By construction, the transferred model has been exposed to more data in total than the model trained from scratch. This is the intended mechanism rather than a confounding one, since the benefit of a pre-trained model is precisely that it arrives having already learned from other recordings.
Both datasets were recorded with the same aerolysin pore, so these results establish transfer across analytes rather than across pore types; whether the learned features generalize to structurally different nanopores remains to be tested.

\section{Discussion}
This work demonstrates that the presented multi-modal transformer architecture, trained to fuse several representations of the same nanopore blockade signal, substantially improves classification performance over other state-of-the-art methods.
Our model identifies 92.6\% of events in a 42-class peptide-ladder data set, exceeding the best single-representation method by more than ten percentage points.
With an estimated 3-5\% of training labels incorrect, this performance approaches the ceiling attainable on this data set.
Further comparison of the per-class accuracy distributions shows that the model is more uniform across classes than any of the alternatives, with a substantially narrower spread and an increase of almost twenty percentage points in the worst-resolved class.
This matters because the reliability of a sensing platform is determined by the analyte it resolves least well, rather than by its average, and a classifier that is uniformly accurate across classes is qualitatively more useful than one that is accurate on average.
Our model achieves both, combining the highest overall accuracy with the narrowest spread across classes.\\

\noindent Our attention analysis offers a partial explanation for the performance gain.
The time-series and wavelet models attend to different structures within the same event, with the former concentrating on the entry and exit phases and on deep, short blockades, and the latter attending throughout the high-frequency band.
The fact that the two representations emphasize different regions is consistent with our observation that no single representation is best for every peptide class.\\

\noindent The architecture places no restriction on the number or type of input branches, so newly identified representations can be integrated without modification.
Additionally, the inputs are not restricted to transformations of a single measurement but could also originate from additional measurements.
The electro-optical readout of the pore, in which optical measurements accompany ionic current recordings~\citep{yang22a, fried22a}, is a natural candidate for joint analysis of this kind, and one our architecture supports as it stands.\\

\noindent Our transfer-learning experiment suggests that the features that the model learns are not specific to a single analyte set.
A model pre-trained on the peptide-ladder dataset and fine-tuned on the $\mathrm{X_{R7}}$ dataset converges faster and attains more uniform per-class accuracy than a model trained from scratch, indicating that it arrives already able to extract relevant signal structure.
Labeled nanopore data are costly to generate, so this reduces the barrier to applying machine learning to new sensing tasks.
Additionally, the results on $\mathrm{X_{R7}}$ show that the model resolves all twenty amino acids with 99.0\% macro-average accuracy, whereas Ouldali et al. reported that only 13 of them can be distinguished by mean residual current alone, the remaining seven falling into two indistinguishable groups~\citep{ouldali20a}.
This confirms that the temporal and spectral structure of a blockade carries information that single-feature analyzes discard.\\

\noindent Two main limitations bound what these results establish.
Both datasets were recorded with the same aerolysin pore, so we demonstrate transfer across analytes rather than across pore architectures.
In addition, 42 peptide classes, while a substantial advance on previous work, remain a small fraction of the diversity a clinical assay would encounter.
Consequently, future work should focus on validation on larger datasets with clean labels and application to data measured with structurally different pores.
If these tests are successful, the architecture could serve as a nanopore-agnostic foundation model for the field.

\section{Materials and data}\label{sec:methods_data}
\subsection*{Experimental data and model input}
\begin{figure}[h]
    \centering
    \begin{minipage}{0.49\linewidth}
        \centering
        \includegraphics[width=1.\linewidth]{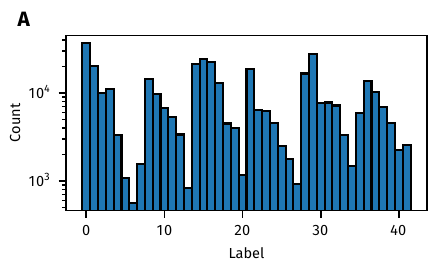}
    \end{minipage}
    \hfill
    \begin{minipage}{0.49\linewidth}
        \centering
        \includegraphics[width=1.\linewidth]{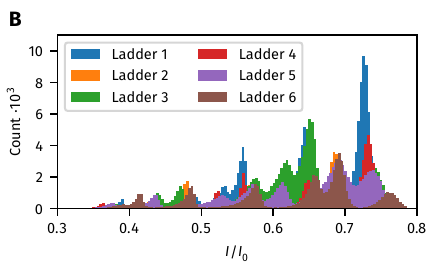}
    \end{minipage}

    \vspace{0.1em}

    \begin{minipage}{0.49\linewidth}
        \centering
        \includegraphics[width=1.\linewidth]{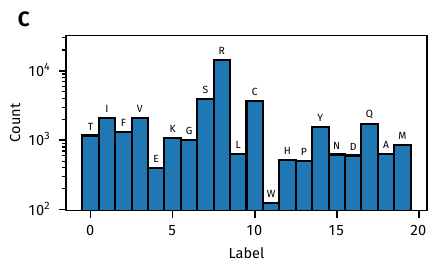}
    \end{minipage}
    \hfill
    \begin{minipage}{0.49\linewidth}
        \centering
        \includegraphics[width=1.\linewidth]{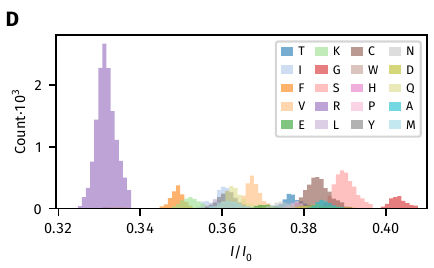}
    \end{minipage}
    \caption{\textbf{Properties of the two datasets.} Top row, peptide-ladder dataset; bottom row, $\mathrm{X_{R7}}$ dataset. (A) Event counts per class for the peptide ladder, highlighting the strong class imbalance. (B) Mean-current histograms for all six peptide ladders. (C) Event counts per class for the $\mathrm{X_{R7}}$ dataset. (D) Mean-current histograms for all 20 amino-acid peptides.}
    \label{fig:dataset}
\end{figure}

We evaluate our model using the peptide ladder data set introduced in~\cite{tovey25b-pre, behrends22a, hossbachPeptideClassificationStatistical2024} and the $\mathrm{X_{R7}}$ data set introduced in~\cite{ouldali20a}. 
The data set for the peptide ladder contains peptides of seven different lengths and six unique amino acid sequences, resulting in 42 distinct classes. 
A list of how the labels correspond to the amino acid sequences is provided in Sec.~\ref{app:LabelList}.
In total, the data set contains 350,000 labeled events, which we divide into 70\% training, 15\% validation, and 15\% test sets. 
Approximately 3--5\% of these events are mislabeled due to limitations in the labeling process, which places an upper bound on the accuracy any classifier can achieve in this dataset.
Furthermore, as shown in Fig.~\ref{fig:dataset} a), events are not distributed equally between the 42 classes. 
Due to experimental limitations, the data inherit a significant class imbalance, posing additional challenges for training. 
Fig.~\ref{fig:dataset} b) shows the histograms over the mean currents of all 42 peptides. 
The figure highlights the difficulty of distinguishing between peptides with similar mass, which produce signals with similar mean currents.

\noindent The data set $\mathrm{X_{R7}}$ contains the residual current events of all 20 proteinogenic amino acids linked to the arginine heptapeptide carriers, forming 20 distinct classes.
This dataset contains 38,000 labeled events across the 20 classes, roughly an order of magnitude fewer than the peptide-ladder dataset.
As for the peptide ladder, we use a split of 70\% / 15\% / 15\%  for training / validation / test.
This data set also exhibits a pronounced class imbalance (Fig.~\ref{fig:dataset} c), and the mean current histograms of the 20 amino-acid peptides show substantial overlap between classes (Fig.~\ref{fig:dataset} d), again illustrating the difficulty of distinguishing peptides based on mean current alone.

\noindent For each of the events in the introduced datasets, we use the time series signal, a wavelet transformation, and the $catch22$~\cite{lubbaCatch22CAnonicalTimeseries2019} feature set as input to our multimodal model. 
Wavelet images are created by performing the integration
\begin{equation}
W(a, b) = \frac{1}{\sqrt{|a|}} \int_{-\infty}^{\infty} x(t) \, \psi^*\left(\frac{t-b}{a}\right) dt,
\label{eqn:mother}
\end{equation}
where $\psi(x)$ is the mother wavelet, $x(t)$ is the time series signal, and $a$ and $b$ are the frequency and time parameters. 
The result can be represented as an image in which each pixel corresponds to the result for a pair of parameters $(a,b)$. 
In this work, the \textit{ssqueezepy} Python package~\cite{muradeli20a} was used to perform the wavelet transformation. 
In line with previous work ~\cite{tovey25b-pre}, the \textit{hhhat} wavelet with parameter $\mu=5$ was employed. 
The resulting output is a two-channel image with real and imaginary parts of the solution. To highlight smaller features, we compute the logarithm of the resulting image and then use the absolute value for each pixel to create the final one-channel input image for our model.
The \textit{catch22}~\cite{lubbaCatch22CAnonicalTimeseries2019} feature set, originally derived for use in data mining, contains 22 features based on different properties of a time series signal. 
In~\cite{hossbachPeptideClassificationStatistical2024}, this set of characteristics was combined with the mean current, the standard deviation, and the natural logarithm of the dwell time of nanopore blockade events. 
To remove redundant data, we reduced this set to the five most relevant features identified in~\cite{hossbachPeptideClassificationStatistical2024} by performing SHAP analysis on the same peptide ladder dataset used in this work.  
These five features form the third input to our model.

\subsection*{Model Design}\label{subsec:architecture}
Our proposed architecture integrates information from various modalities, including time series, image, and feature sets. 
To achieve this, the input is divided into different branches, each corresponding to a specific modality. 
These branches are processed and correlated in the multi-branch transformer encoder, as illustrated in Figure~\ref{fig:architecture}. 
Subsequently, the output is used for classification.
Before passing the different modalities to the multi-branch transformer encoder, they are projected into the model's latent space. 
For image inputs, we adopt the procedure described in~\cite{ImageWorth16x16}. 
The image is divided into quadratic patches, which are flattened and projected in a space of dimensionality $D_{3}$ with a linear layer. 
To capture the spatial relationships between these patches, a learnable positional embedding is added. 
In addition, a classification token (C\textsubscript{3}) is concatenated into the sequence to collect relevant information for classification. 
The final sequence of tokens forms branch two.
Applying this concept to time series data, as shown on the left in Figure~\ref{fig:architecture}, is straightforward. 
The time series is divided into sections of equal length and projected into a space of dimensionality $D_{1}$ using a linear layer. 
For the time series input, a learnable positional embedding is added to capture temporal correlations within the signal, and a classification token (C\textsubscript{1}) is concatenated into the sequence, forming the branch one.
Incorporating the feature set, shown in the middle of Figure~\ref{fig:architecture}, requires a different approach, since transformers typically do not handle such data directly. 
For this modality, we generate another classification token (C\textsubscript{2}) without additional tokens in its branch. 
This is achieved by passing the descriptor through two linear layers with a GELU activation function between them, projecting it into the dimension $D_{2}$. 
We use multiple layers and introduce nonlinearity to create C\textsubscript{2} as it will not be further processed in the multi-branch transformer encoder.
To transfer descriptor information into the two other branches, we use the cross-attention mechanism as introduced in~\cite{chenCrossViTCrossAttentionMultiScale2021} once before they enter the multi-branch transformer encoder. 
To achieve this, we use a linear transformation to project the tokens $C_{1/3}$ into the same dimension as $C_{2}$, perform multi-head attention, and then transform $C'_{1/3}$ back to their original dimension to get $C^{\text{new}}_{1/3}$. 
For $C_{1}$, this process can be expressed as
\begin{equation}
    \begin{gathered}
        C'_{1} = f^{12}(C_{1}), \quad q = C'_{1}W_{q}, \quad k = C_{2}W_{k}, \quad v = C'_{1}W_{v}\\
        C^{\text{new}}_{1} = \text{norm}\left[C_{1} + f^{21}\left(\text{softmax}\left(qk^{T}/\sqrt{D_{2}/h}\right)v\right)\right],
    \end{gathered}
\label{eq:1}
\end{equation}
where $h$ is the number of heads, $W_{q}$, $W_{k}$ and $W_{v}$ are learnable parameters and $f^{12}$ and $f^{21}$ are learnable projections to adjust dimensions. 
This is also done for $C_{3}$. 
Only $C_{2}$ does not take information from the other classification tokens, as they do not yet hold any information about their branches.
After the described preparation, branches one and two are passed into the multi-branch transformer encoder. 
This encoder operates in two steps: the first step processes the information within each branch independently and the second step mixes the information between different branches. 
These two steps are repeated N times to process the information. 

We use the standard transformer encoder block introduced in~\cite{vaswaniAttentionAllYou2017} to update the branches individually.
After the first transformer encoder block, cross-attention spreads information between branches. 
The central difference from the description in Equation~\ref{eq:1} is that the key $k$ is not created using only the classification token, but consists of the entire sequence of one of the branches. 
As a result, during this step, the classification token $C_{1}$ receives information about the whole sequence in branch two and vice versa. 
Using cross-attention instead of all attention to correlate information between branches drastically reduces the number of operations done. 
Moreover, since each branch contains information from different modalities, it is more efficient for the transformer to process them in separate latent spaces.
After the encoder block, we pass the three classification tokens through separate multilayer perceptrons (MLPs) and combine their outputs for the final classification.
The described architecture is designed to incorporate as many modalities as possible while maintaining minimal computational overhead by utilizing the cross-attention mechanism. 
In addition, the dimensionality, the number of tokens, and the number of transformer encoder blocks can be chosen individually for each branch. 
This flexibility allows for the fine-tuning of the number of parameters and computational operations in each branch according to the complexity of the information being processed.
The exact parameters of the architecture used in this study are presented in Sec.~\ref{app:ModelParams}.

\subsection*{Model training}
\begin{figure}[h]
    \centering
    \includegraphics[width=0.7\linewidth]{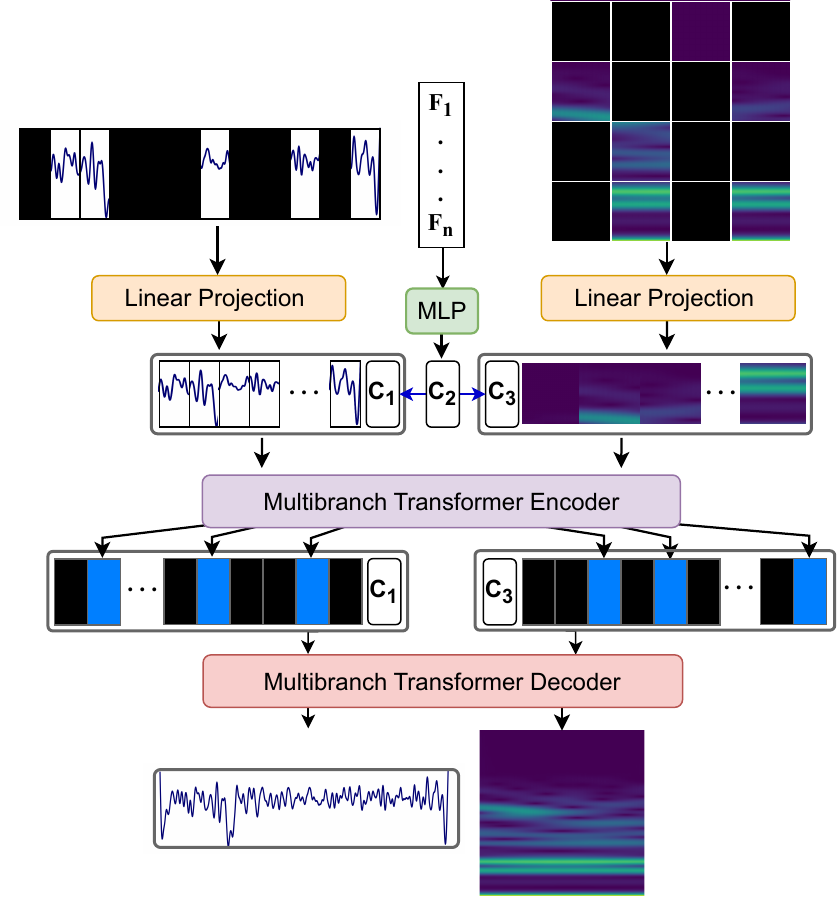}
    \caption{\textbf{Pre-training Framework.} During pretraining, the model is trained to reconstruct time series and image data from small portions of unmasked input tokens. This is achieved using an encoder-decoder architecture. The encoder processes the unmasked patches to generate a latent space representation, which is then concatenated with the masked patches in their original order. The decoder takes this combined input and attempts to reconstruct the original time series signals and wavelet images. In our approach, 50\% of the time series tokens and 70\% of the wavelet image tokens are masked during pretraining.}
    \label{fig:pretraining}
\end{figure}

Model training is divided into two stages: self-supervised pretraining followed by supervised fine-tuning of the classification task. 
Previous work has shown that pretraining can accelerate convergence and improve downstream accuracy, particularly in small data sets where it enhances the performance of transformer-based models~\cite{devlin19a, erhan10a, PretrainingLightweightVision2024}.
As illustrated in Fig.~\ref{fig:pretraining}, our pretraining procedure follows the masked autoencoding strategy introduced in~\cite{heMaskedAutoencodersAre2022}. 
During this phase, random tokens from both the time series and wavelet images are masked and replaced with separate learnable mask tokens. 
Only unmasked tokens are passed to the multi-branch transformer encoder. 
The encoded output is then recombined with the mask tokens in their original positions and passed through a lightweight decoder to reconstruct the original input. 
This encoder-decoder setup reduces computational overhead while maintaining reconstruction performance. 
The masking ratio can be set individually per branch, depending on the information content.
A key modification to the original method is the inclusion of classification tokens during the pretraining process. 
Since different parts of the input are masked in each modality, this encourages the model to share information across branches, thereby improving its ability to reconstruct the full input and correlate information between branches for classification. 
The same principle applies to incorporating descriptor data at this stage.
Given the experimental difficulty in obtaining large volumes of accurately labeled data, the ability to leverage unlabeled data through pretraining would represent a substantial practical advantage.
Similar approaches in other research fields have been shown to benefit greatly from pre-training on large amounts of unlabeled data~\cite{xu_limu-bert_2021, yuan_self-supervised_2021}.\\
For the transfer-learning dataset, we train one model as described above, including pre-training and fine-tuning on the $\mathrm{X_{R7}}$ dataset.
Additionally, the transferred model was first trained on the peptide-ladder dataset in both pre-training and supervised fine-tuning.
Subsequently, the classification head was removed and replaced with a newly initialized classification head with 20 output dimensions, while all remaining weights were retained and left trainable.
The model was then fine-tuned on the $\mathrm{X_{R7}}$ dataset with the same learning rate schedule and augmentation as the one fine-tuned on the peptide-ladder dataset.
No additional pre-training was performed on the $\mathrm{X_{R7}}$ dataset for the transferred model.


\newpage

\appendix
\section{Label list}\label{app:LabelList}
\begin{table}[h]
    \centering
    \caption{Label used in the classifier and the corresponding peptide.}
    \begin{tabular}{|c|l|l||c|l|l|}
        \hline
        Label & Peptide & Sequence & Label & Peptide & Sequence \\
        \hline
        0  & L1AS4  & H-RRRR-OH       & 21 & L4AS4  & H-KRRR-OH        \\
        1  & L1AS5  & H-YRRRR-OH      & 22 & L4AS5  & H-SKRRR-OH       \\
        2  & L1AS6  & H-KYRRRR-OH     & 23 & L4AS6  & H-ASKRRR-OH      \\
        3  & L1AS7  & H-SKYRRRR-OH    & 24 & L4AS7  & H-RASKRRR-OH     \\
        4  & L1AS8  & H-ASKYRRRR-OH   & 25 & L4AS8  & H-SRASKRRR-OH    \\
        5  & L1AS9  & H-RASKYRRRR-OH  & 26 & L4AS9  & H-YSRASKRRR-OH   \\
        6  & L1AS10 & H-SRASKYRRRR-OH & 27 & L4AS10 & H-RYSRASKRRR-OH  \\
        \hline
        7  & L2AS4  & H-SRRR-OH       & 28 & L5AS4  & H-YRRR-OH        \\
        8  & L2AS5  & H-RSRRR-OH      & 29 & L5AS5  & H-AYRRR-OH       \\
        9  & L2AS6  & H-ARSRRR-OH     & 30 & L5AS6  & H-RAYRRR-OH      \\
        10 & L2AS7  & H-YARSRRR-OH    & 31 & L5AS7  & H-SRAYRRR-OH     \\
        11 & L2AS8  & H-RYARSRRR-OH   & 32 & L5AS8  & H-SSRAYRRR-OH    \\
        12 & L2AS9  & H-SRYARSRRR-OH  & 33 & L5AS9  & H-RSSRAYRRR-OH   \\
        13 & L2AS10 & H-KSRYARSRRR-OH & 34 & L5AS10 & H-KRSSRAYRRR-OH  \\
        \hline
        14 & L3AS4  & H-YRRR-OH       & 35 & L6AS4  & H-ARRR-OH        \\
        15 & L3AS5  & H-RYRRR-OH      & 36 & L6AS5  & H-RARRR-OH       \\
        16 & L3AS6  & H-SRYRRR-OH     & 37 & L6AS6  & H-SRARRR-OH      \\
        17 & L3AS7  & H-ASRYRRR-OH    & 38 & L6AS7  & H-YSRARRR-OH     \\
        18 & L3AS8  & H-RASRYRRR-OH   & 39 & L6AS8  & H-RYSRARRR-OH    \\
        19 & L3AS9  & H-SRASRYRRR-OH  & 40 & L6AS9  & H-KRYSRARRR-OH   \\
        20 & L3AS10 & H-KSRASRYRRR-OH & 41 & L6AS10 & H-SKRYSRARRR-OH \\
        \hline
    \end{tabular}
\end{table}
\label{app:theorem}

\section{Extended attention analysis}\label{app:AttAnal}

The following section extends the attention analysis of the main text, resolving the final-layer attention by individual heads and showing attention rollout for additional example signals.

\begin{figure}[h]
    \centering
    \includegraphics[width=0.99\linewidth]{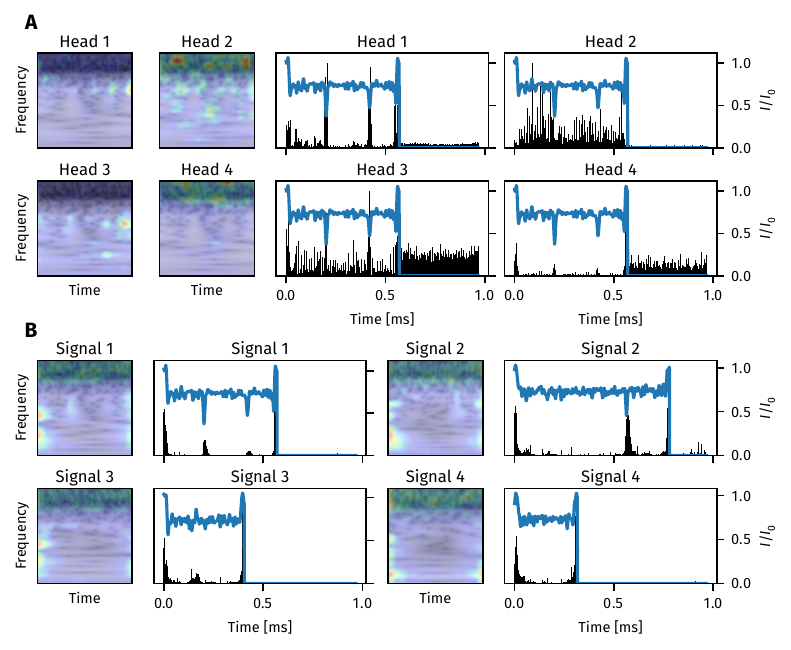}
    \caption{\textbf{Extended attention analysis of the vision transformer and the time series transformer.} For the vision transformer, the wavelet images are displayed in grayscale, with attention weights shown in the foreground, where red represents the highest values. For the time series transformer, the attention weights are displayed as black bars, with higher bars representing higher attention weights. Since the zero padding to a uniform input length is not masked in the transformer, this part is shown as well. a) The attention weights in the last transformer layer corresponding to the classification token are displayed for the four different attention heads of one example signal. b) Attention rollout~\cite{abnar_quantifying_2020} is displayed for four example signals. Signal one corresponds to the signal in a).}
    \label{fig:app:attention}
\end{figure}

\noindent Fig.~\ref{fig:app:attention}A shows attention weights from the final transformer layer associated with the classification token, displayed separately for each of the four attention heads.
Since this token aggregates information for the final prediction, its attention weights indicate which parts of the input are most influential, and the per-head decomposition reveals that the heads attend to different aspects of the input in parallel.

\noindent For the time series model, attention is focused on deep blockades and on the entry and exit phases of the peptide signal.
Some heads also exhibit increased attention to the padded signal segments.
This may indicate that the model uses the empty region as a reference frame to infer dwell time or that it uses it to correlate information extracted from the signal itself, as observed in~\citep{darcet24a-pre}.

\noindent For the wavelet representation, the heads do not primarily localize distinct time regions but instead emphasize different frequency bands, some focusing on high-frequency components and others attending more strongly to mid- and low-frequency regions.
This spectral separation indicates that the model isolates complementary frequency-domain features.
The focus on high-frequency regions is consistent with previous findings that these components contain meaningful information rather than pure noise~\citep{tovey25b-pre}.

\noindent Fig.~\ref{fig:app:attention}B shows attention rollout~\citep{abnar_quantifying_2020} for four example signals, the first of which corresponds to the signal in panel A.
Across these signals, both models consistently highlight the entry and exit phases as well as deep, short blockades, supporting the observations from the final-layer attention.
 Deep, short blockades have been hypothesized to arise from a binding site near the cis opening of the pore, where a positively charged physico-electric barrier and a reduced volume are located~\citep{ensslen24b, boukhet18a}, suggesting that peptide interactions with this site may provide discriminative information for classification.
We inspected attention patterns across other randomly selected signals and did not observe qualitatively different behavior.

\section{Model parameters}\label{app:ModelParams}
The multi-modal transformer encoder used for our experiments consists of 12 transformer encoder blocks for branches one and two, with cross-attention applied after every encoder block. For both the time series and the wavelet image branch, we use a dimension of $D_{1}, D_{3} = 256$, a hidden dimension of 1024 and eight attention heads. Tokenization is performed using a sequence size of 4 for time series data and a patch size of 14 for wavelet images. The descriptor data are passed through an MLP of architecture ($D^{128}\cdot\text{GELU}\cdot D^{256}$). After the encoder block, all classification tokens are processed by separate MLPs of architecture ($D^{128}\cdot\text{GELU}\cdot D^{42}$). These specifications add up to roughly 28M learnable parameters in our model.\\
The raw time series transformer we use has 12 transformer encoder blocks of dimension 256, each with four attention heads. The tokenization and classification follow the same procedure as the time series branch in our multi-modal model. The time series transformer adds up to 9.5M learnable parameters.

\section{Model training specifications}
During pre-training of the multi-modal model, 50\% of the time series tokens and 70\% of the wavelet image tokens are masked. Training takes place with a batch size of 1600, using the MSE error function. The learning rate starts with a linear warm-up to a maximum of $8\cdot 10^{-5}$, followed by a cosine decay \cite{loshchilovSGDRStochasticGradient2017}. To make the model more robust, we apply augmentation and regularization techniques. During pre-training, we use dropout and weight decay for regularization. Data augmentation consists of random flips (reversing time) and random cutting for the raw time series, along with random horizontal flips and random resize cropping for the wavelet data.\\
For classification, we use cross-entropy loss, a batch size of 800 and the same learning rate schedule as for pre-training with a maximum learning rate of $4\cdot 10^{-5}$. The same augmentation and regularization strategies from pre-training are applied during classification. Additionally, we incorporate label smoothing and apply a random constant shift to the time series data.\\
The time series transformer follows the same pretraining and classification procedure as the time series branch in the multi-modal model. The same batch sizes, learning rate schedules, and peak learning rates are used.\\
All specifications above apply to both the peptide-ladder and the $\mathrm{X_{R7}}$ datasets.

\section*{Acknowledgments}
The authors acknowledge Michel Mom for rendering the final version of the nanopore figure.
The authors thank Tobias Ensslen and Jan C. Behrends for providing the experimental data of the peptide-ladder dataset.
\paragraph*{Funding:}
This project was funded by the BMBFTR through the nanodiagBW project, grant No. 03ZU1208AM.
The authors acknowledge financial support from the German Funding Agency (Deutsche Forschungsgemeinschaft DFG) under Germany's Excellence Strategy EXC 2075-390740016.
This work was supported by SPP 2363- "Utilization and Development of Machine Learning for Molecular Applications – Molecular Machine Learning."
Funded by the Deutsche Forschungsgemeinschaft (DFG, German Research Foundation), Project-No 497249646.\\
Computations were performed on the ICP Compute Cluster, Grant No. INST 41/1148-1 Holm 492175459, and through the INST 35/1597-1 FUGG grant. \\
\paragraph*{Author contributions:}
S.K. proposed the multi-modal fusion approach, developed and implemented the multi-modal architecture, conducted model training and analysis, drafted parts of the manuscript and performed editing.
J.H. curated the experimental datasets into a machine-learning ready format, drafted parts of the manuscript and performed editing.
S.T. proposed the use of transformer architectures for the project, supervised the machine-learning development, contributed to study design, drafted parts of the manuscript and performed editing.
C.H. supervised the investigation and provided feedback. All authors read
the manuscript and provided editing feedback.

\paragraph*{Competing interests:}
The authors declare the following competing financial interests: The University of Stuttgart
has filed a patent application with patent number DE 10 2025 123 407.8 for the methods
introduced in this work.
\paragraph*{Data, code and materials availability:}
Upon publication, the data will be made available upon reasonable request to the relevant authors. 
Upon publication, all scripts used will be published to https://doi.org/10.18419/DARUS-6326.

\vskip 0.2in
\bibliographystyle{plain} 
\bibliography{references}

\end{document}